\documentclass{article} 
\usepackage{iclr2021/iclr2021_conference,times}
\usepackage{mathtools}
\usepackage{enumerate}

\usepackage{caption}
\usepackage{amsthm}

\usepackage{subcaption}
\usepackage{graphicx}

\usepackage{natbib}

\usepackage{amsfonts,amsmath,amssymb}
\usepackage{bbm}
\usepackage{xspace}

\usepackage{multicol}
\usepackage{enumitem}

\providecommand{\mc}[1]{\mathcal{#1}}

\usepackage{algorithm}
\usepackage{algpseudocode}

\usepackage{wrapfig}

\usepackage{subcaption}

\usepackage{hyperref}
\usepackage{url}

\title{Tracking the perspectives of\\ interacting language models}

\author{Hayden Helm$^{\dagger}$ \\ Nomic AI 
\And 
Brandon Duderstadt \\
Nomic AI
\And 
Youngser Park\\
Center for Imaging Sciences \\
Johns Hopkins University
\And
Carey E. Priebe \\
Dept. of Applied Math. \& Statistics \\
Johns Hopkins University
}

\newcommand\blfootnote[1]{%
  \begingroup
  \renewcommand\thefootnote{}\footnote{#1}%
  \addtocounter{footnote}{-1}%
  \endgroup
}

\iclrfinalcopy 
\begin{document}
\maketitle
\blfootnote{\\
$ ^{\dagger} $ corresponding author; \texttt{[first-name]}@nomic.ai 
}

\vspace{-1.25cm}

\begin{abstract}
Large language models (LLMs) are capable of producing high quality information at unprecedented rates.
As these models continue to entrench themselves in society, the content they produce will become increasingly pervasive in databases that are, in turn, incorporated into the pre-training data, fine-tuning data, retrieval data, etc. of other language models.
In this paper we formalize the idea of a communication network of LLMs and introduce a method for representing the perspective of individual models within a collection of LLMs. 
Given these tools we systematically study information diffusion in the communication network of LLMs in various simulated settings. 
\end{abstract}

%




\begin{figure*}[h!]
\centering
\captionsetup[subfigure]{justification=centering}
\begin{subfigure}{0.24\textwidth}
    \centering
    \includegraphics[width=\linewidth]{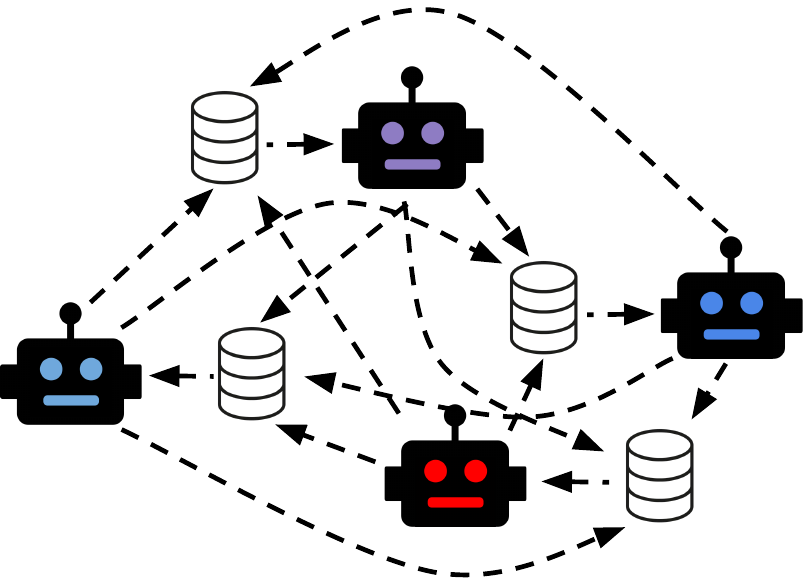}
    \subcaption{Fully connected.}
\end{subfigure}
\begin{subfigure}{0.24\textwidth}
    \centering
    \includegraphics[width=\linewidth]{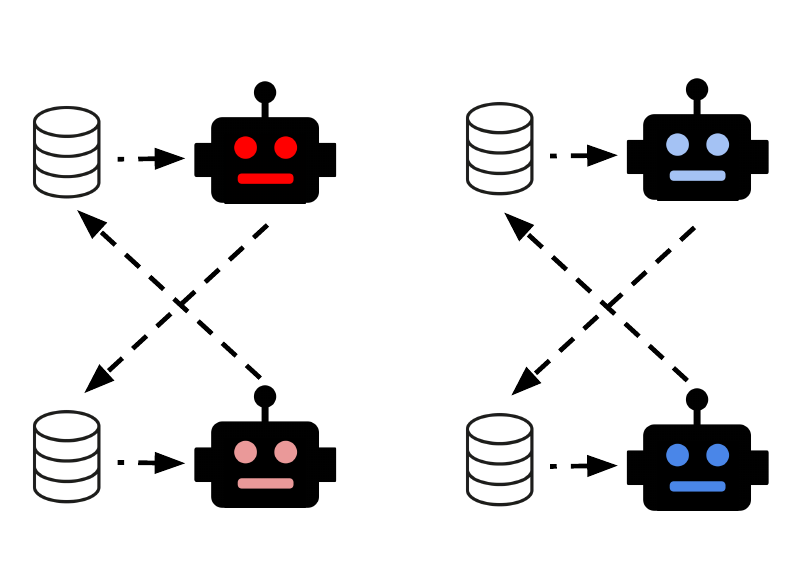}
    \subcaption{Intra-class only.}
\end{subfigure}
\begin{subfigure}{0.24\textwidth}
    \centering
    \includegraphics[width=\linewidth]{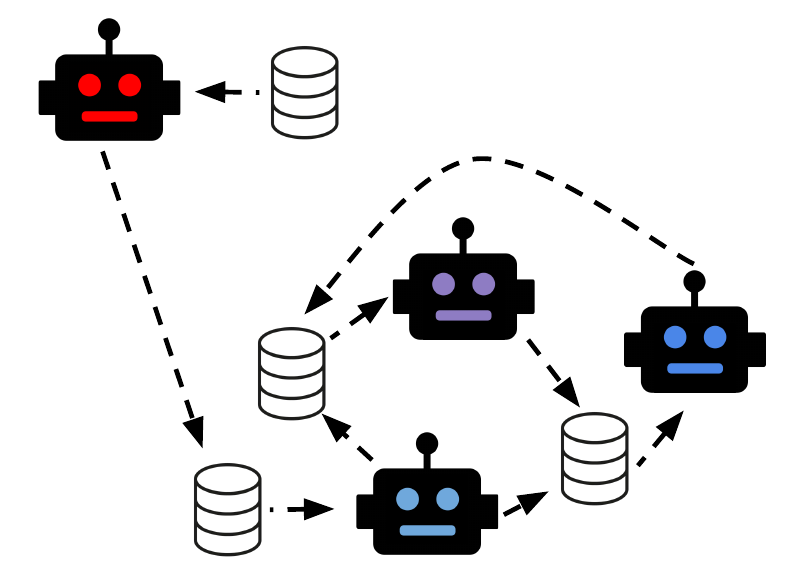}
    \subcaption{Vulnerable.}
\end{subfigure}
\begin{subfigure}{0.24\textwidth}
    \centering
    \includegraphics[width=\linewidth]{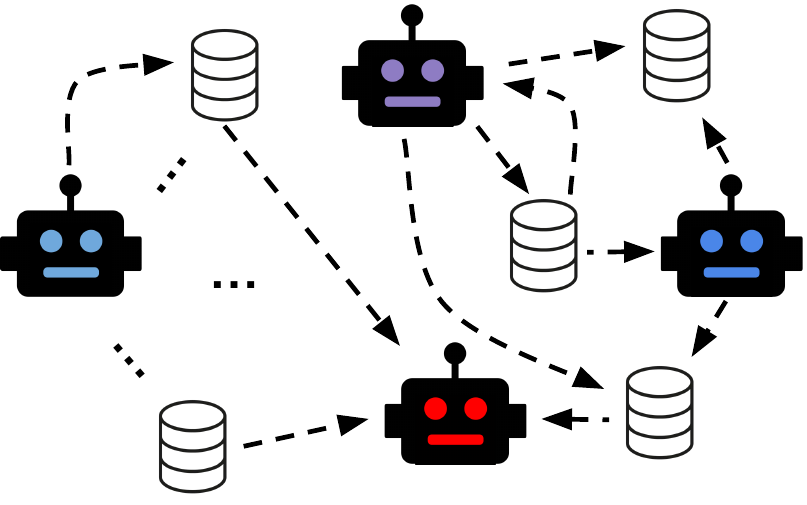}
    \subcaption{General.}
\end{subfigure}
\caption{Examples of communication networks of language models and databases. The edge structure and model intitializations directly impact the evolution of the perspectives of the models and the overall health of the system.}
\label{fig:network-examples}
\end{figure*}

\section{Introduction}

The success of large pre-trained models in natural language processing \citep{devlin2018bert}, computer vision \citep{oquab2023dinov2}, signal processing \citep{radford2023robust}, among other domains \citep{jumper2021highly} across various computing and human benchmarks has brought them to the forefront of the technology-centric world. 
Given their ability to produce human-expert level responses for a large set of knowledge-based questions \citep{touvron2023llama, achiam2023gpt}, the content they produce is often propagated throughout forums that have influence over other models and human users \citep{Brinkmann_2023}. As such, it is important to develop sufficient frameworks and complementary tools to understand how information produced by these models affects the behavior of other models and human users. We refer to a system where a model can potentially influence other models as a system of interacting language models.

Beyond their ability to influence information on human-model forums, systems of interacting language models are interesting in their own right --
insofar as an individual model is an intriguing proxy for an individual human \citep{helm2023statistical, kosinski2023evaluating}, a system of interacting language models is an intriguing proxy for human communities.
Systems of interacting language models are thus an alluring alternative or complement to studying human communities in the social sciences.
For example, it is often infeasible or unethical to subject entire communities to different information paradigms to understand how individuals within the community -- as well as the community itself -- change in response to an intervention.
These issues are less prominent for systems of interacting language models.
Further, there is potential for greater control in community membership and cross-community interactions, which may improve reproducibility and mitigate the effects of sociological confounders.




In this paper, we study information diffusion in a system of interacting language models. 
The framework and methods that we develop can be applied to monitoring information diffusion in human-model forums and to the treatment of systems of interacting language models quantitatively as proxy human communities.
The current standard \citep{perez2024cultural} for studying information diffusion in a system of interacting language models requires i) parameterizing models with different system prompts, contexts, weights, or collections of data, ii) providing an environment or template for model-to-model or model-to-dataset interactions, and iii) analyzing how the outputs of the models change after a sequence of interactions.

For example, researchers include descriptions of desired model behavior or personality in the system prompt -- e.g., ``You have opinion \textit{A}" is included in the system prompt for model 1 and ``You have opinion \textit{B}" is included in the system prompt for model 2, etc. -- to promote diversity in model response \citep{park2023generative, chuang2023simulating, papachristou2024network}. 
While the intended model response diversity is achieved, previous studies have failed to quantitatively assess the effect of different model initializations and, instead, rely on qualitative checks. 
Similarly, analyzing changes in model responses as the system evolves has previously been limited to human inspection of responses \citep{park2023generative},
or classification of responses into a few classes \citep{chuang2023simulating}.

We introduce the \textit{perspective space} of a collection of models to address the gap in quantitative methods for studying the diversity and evolution of model responses. 
The perspective space is an embedding-based representation of a collection of models designed to capture the relative differences in model responses for a fixed set of prompts. 
The method can be used to study information diffusion and general system dynamics by querying each model with the same set of queries at each time step. 
To demonstrate the effectiveness of the perspective space for understanding model-level diversity and for analyzing model-level and system dynamics, we formalize the system of interacting language models as a graph. 
The formalization enables systematic study of the effect of different communication structures on information diffusion that is otherwise not possible.

Our contribution is two-fold:
i) We model a system of interacting language models as a graph and systematically study the effect of different communication structures on information diffusion. 
ii) We introduce the perspective space as a method to quantitatively analyze information diffustion in a population of language models.





\label{introduction}


\section{A communication network of LLMs}
Consider a system that consists of a collection of language of models $ \mathcal{F} = \{f_{1}, \hdots, f_{n}\} $ and databases $ \mathcal{D} = \{D_{1}, \hdots, D_{n'}\} $. 
Given a set of prompts $ \mathbf{X} $, systems deploying model $ f \in \mathcal{F} $ may use the database $ D \in \mathcal{D} $  -- via fine-tuning, context retrieval, etc. -- to produce more relevant outputs with respect to $ \mathbf{X} $.
The outputs of the updated model may be used to update a (potentially different) database $ D' \in \mathcal{D} $. The updated database can then be used as a fine-tuning, retrieval, etc. database for a (potentially different) model $ f' \in \mc{F} $. 
This set of interactions between a model and a database may occur across various models and various databases in the system.

As described, this system can be modeled as a graph $ G=(V, E) $ where $ V = \mathcal{F} \cup \mathcal{D} $ and the directed edge $ (v, v') $ is in $ E $ if vertex $ v $ has influence on vertex $ v' $.
For example, the edge $ (D, f) $ exists if $ f $ has access to $ D $ for retrieval augmentation or if it can use a subset of $ D $ as fine-tuning data. 
Conversely, the edge $ (f, D) $ exists if the output of $ f $ can influence the content of dataset $ D $. 


Our primary interest is the dynamics of a system of interacting LLMs and databases where the vertex and edge sets are indexed by a discrete variable $ t \in \{1, \hdots, T\} $. 
There are many ways components of the graph may vary in $ t $  in such a system. 
For example, the dataset $ D^{(t)} \in V^{(t)} $ may be updated based on the outputs of the model $ 
f^{(t)} \in V^{(t)} $ or the model $ f^{(t)} $ may change after fine-tuning on the contents of the dataset $ D^{(t)} $. 
In both cases $ V^{(t)} \neq V^{(t+1)} $. Similarly, external factors such as the terms of use for a dataset may change to disallow its use for retrieval augmentation or a model may lose write-access to a dataset. 
In both cases $ E^{(t)} \neq E^{(t+1)} $. Figure \ref{fig:network-examples} illustrates simple examples of systems of LLMs as graphs, including three structures that are studied in the simulated settings in Section \ref{sec:simulations}.

\section{Defining a perspective space with surrogate data kernels}

The system-of-LLMs-as-a-graph perspective provides a framework to systematically study the effect of different vertex sets and edge structures on the flow of information through the system as a function of $ t $. 
The framework does not, however, provide a method to track the information flow. 
For this, we introduce an adaptation of the embedding-based data kernel presented in \citep{duderstadt2023comparing}.
For our purposes, an embedding function $ g $ is a mapping to real-valued vectors.

\subsection{The data kernel \& its surrogate}
\begin{figure*}[t]
\centering
\captionsetup[subfigure]{justification=centering}
\begin{subfigure}{0.45\textwidth}
    \centering
    \includegraphics[width=\linewidth]{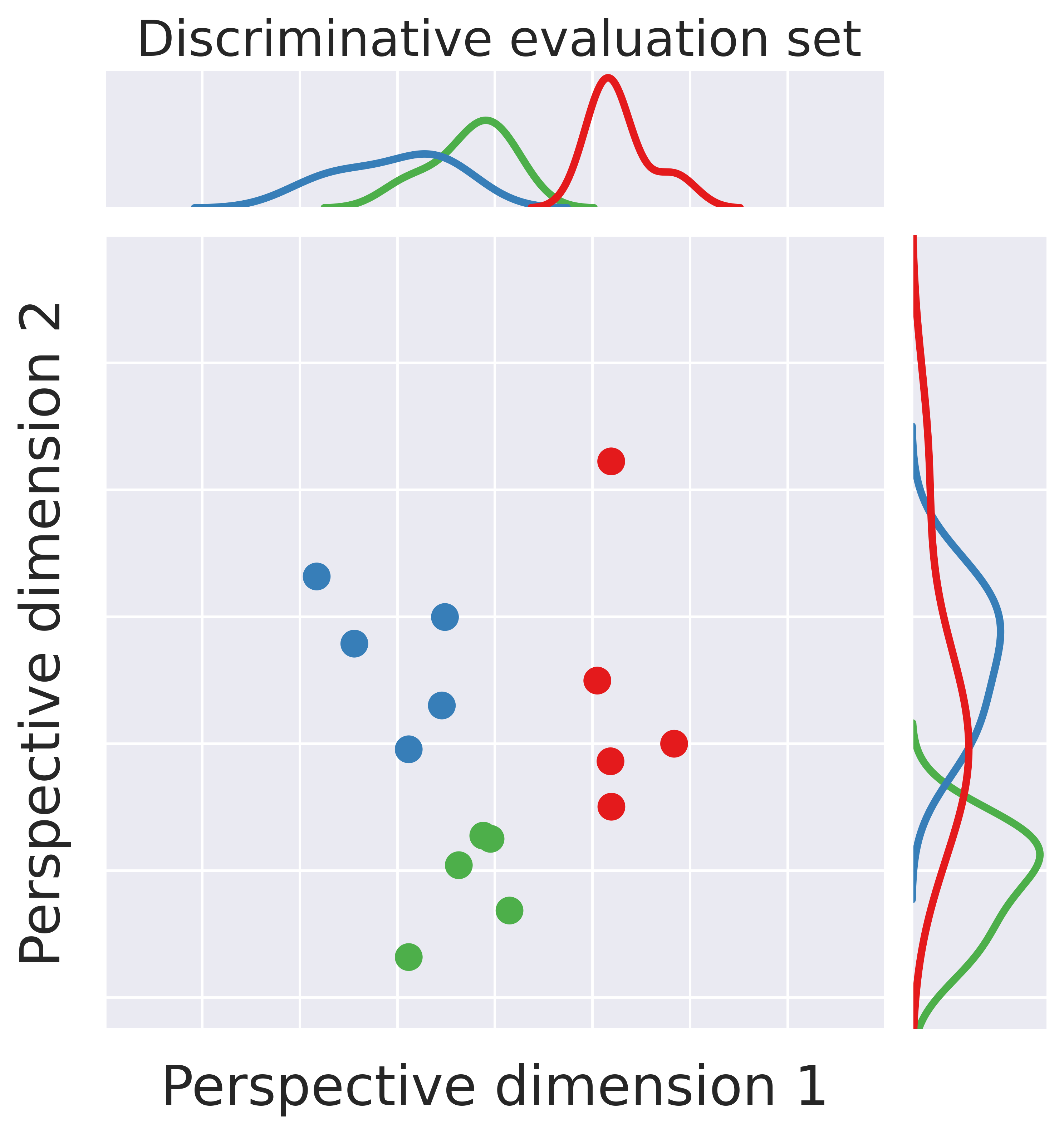}
\end{subfigure}
\begin{subfigure}{0.45\textwidth}
    \centering
    \includegraphics[width=\linewidth]{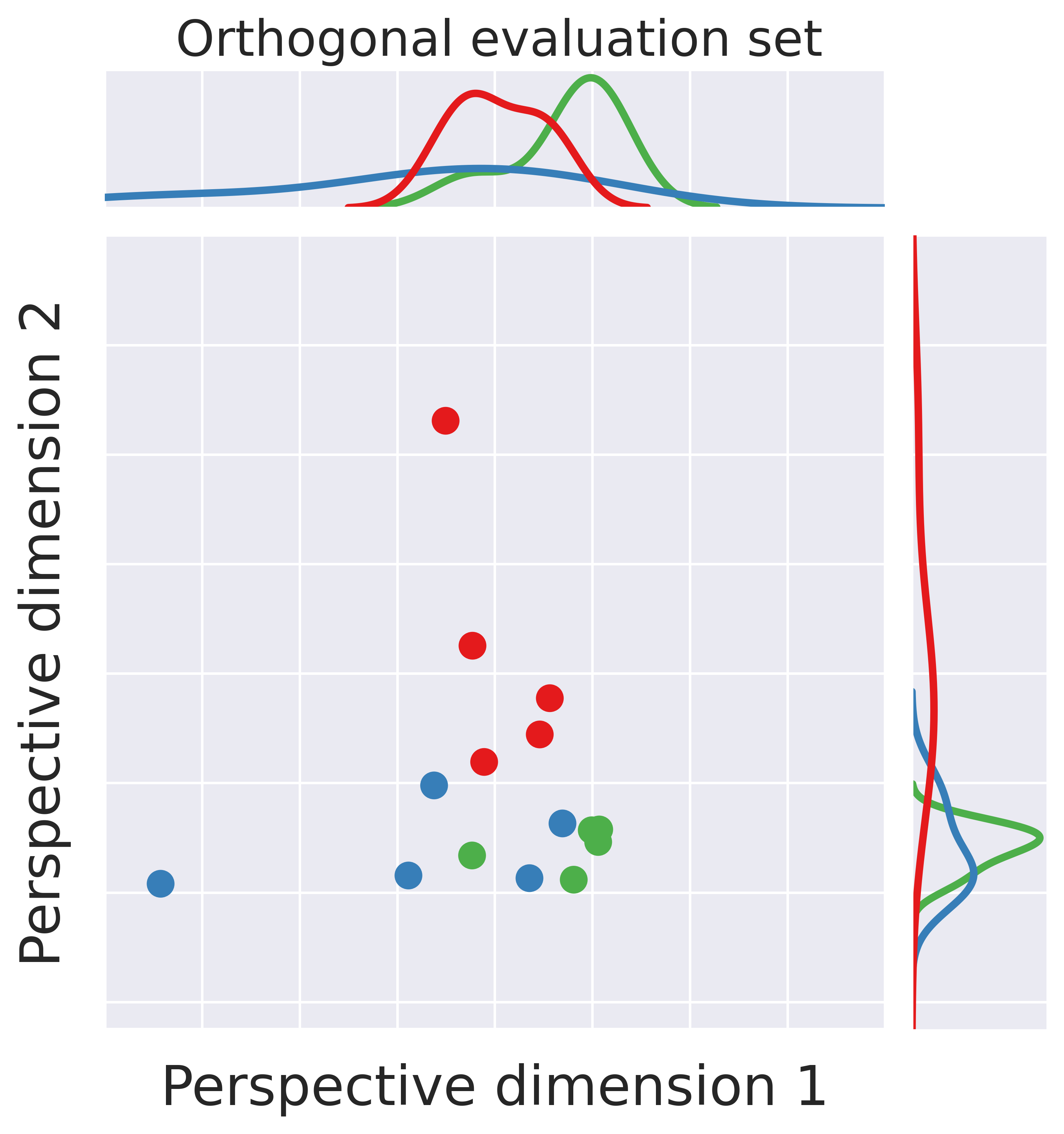}
    \end{subfigure}
\caption{
Two 2-d perspective spaces of fifteen models (5 models each from three classes, encoded by color). 
An evaluation set containing prompts relevant to the differences in the models (left) is better suited to induce a discriminative perspective space than an evaluation set containing ``orthogonal" prompts.}
\label{fig:perspective-space}
\end{figure*}

We let $ \mathbf{X} = \{x_{1}, \hdots, x_{m}\} $ be a collection of prompts with $ x \in \mathcal{X} $ and $ f(\mathbf{X}) = \{f_{\theta}(x_{1}), \hdots, f(x_{m})\} $ be the corresponding set of responses with $ f(x) \in \mathcal{X}' $.
Given an embedding function $ g_{i} $ associated with $ f_{i} $,  recall that the data kernel $ A(g_{i}, \mathbf{X}) $ of the evaluation dataset $ \mathbf{X} $ from the perspective of $ f_{i} $ captures the intrinsic geometry of the embedding space with respect to $ \mathbf{X} $. 
The data kernel enables datum-level and global comparisons of two models with potentially different architectures, sizes, etc. where direct comparison of $ g_{i}(\mathbf{X}) = [g_{i}(x_{1}), \hdots, g_{i}(x_{m})]^\top \in \mathbb{R}^{m \times p} $ and  $ g_{j}(\mathbf{X}) \in \mathbb{R}^{m \times p'} $ is otherwise not possible.

The methodology can be extended to compare the embedding spaces of multiple models $ f_{1}, \hdots, f_{n} $ at once by considering the pairwise distance matrix of the corresponding data kernels. 
In particular, the classical multi-dimensional scaling \citep{torgerson1952multidimensional}) of the $ n \times n $ matrix M with entries $ M_{ij}~=~||~A(g_{i}, \mathbf{X})~-~A(g_{j}, \mathbf{X})~||_{F} $ yields $ d $-dimensional Euclidean representations of the model $ f_{i} $ with respect to $ \mathbf{X} $.
After this transformation, inference methods designed for Euclidean objects can be used for model-level analysis.

The data kernel, as defined in \citep{duderstadt2023comparing}, requires the model $ f_{i} $ to have an associated embedding function $ g_{i} $.
Unfortunately, for some state-of-the-art LLMs such as OpenAI's GPT series, Anthropic's Claude series, etc., an associated embedding function is unavailable and the data kernel cannot be constructed. 
To rectify this, we replace a model's associated embedding function with a \textit{surrogate} embedding function $ \tilde{g}: \mathcal{X}' \to \mathbb{R}^{p} $ that is not necessarily related to any of the LLMs under study.

The surrogate embedding function is not a drop-and-replace solution for model comparisons, however, since the embedding $ \tilde{g}(\mathbf{X}) $ is independent of $ f_{i} $. 
Instead, we query the model with the elements of $ \mathbf{X} $ and embed the responses $ f_{i}(\mathbf{X}) $ with $ \tilde{g} $: the \textit{surrogate data kernel} $ A\left(\tilde{g}, f_{i}(\mathbf{X})\right) $ is simply $ \tilde{g}\left(f_{i}(\mathbf{X})\right) \in \mathbb{R}^{m \times p}$.


\subsection{The perspective space}

As with the original data kernel, we can use the surrogate data kernel to compare the responses from multiple models simultaneously via the CMDS of the pairwise distance matrix $ \tilde{M} $ with entries $ \tilde{M}_{ij}~=~||\tilde{g}(f_{i}(\mathbf{X}))~-~\tilde{g}(f_{j}(\mathbf{X}))||_{F} $. 
We let $ Z_{i} \in \mathbb{R}^{d} $ denote the $ d $-dimensional vector representation of $ f_{i} $. 

Since the representations $ Z_{1}, \hdots, Z_{n} $ are a function of the differences in the model responses -- or ``perspectives" -- $ f_{1}(\mathbf{X}) , \hdots, f_{n}(\mathbf{X}) $, we refer to the subspace populated by $ \{Z_{1}, \hdots, Z_{n}\} $ as the \textit{perspective space} of $ \mathcal{F} $ with respect to $ \mathbf{X} $. The information that is captured by the perspective space depends on $ \tilde{g} $ and $ \mathbf{X} $. 
In particular, $ \tilde{g} $ needs to be able to distinguish between concepts that are intended to be distinguished. For example, a random mapping from $ \mathcal{X}' $ to $ \mathbb{R}^{p} $ is likely insufficient for comparing models, general-purpose embedding functions \citep{reimers2019sentence,nussbaum2024nomic} should be sufficient for capturing the majority of signal, and domain-specific embedding functions \citep{risch2019domain} should be used when the difference in models is highly nuanced.
Similarly, $ \mathbf{X} $ should contain prompts that the models are expected to have meaningfully different responses.
We demonstrate this in Figure \ref{fig:perspective-space} where $ \tilde{g} $ is fixed, $ \mathcal{F} $ consists of 15 models (5 each from three different classes) and $ \mathbf{X} $ is chosen to be relevant to the difference in classes (left) or ``orthogonal" to the difference in classes (right). 
The perspective space is more discriminative (i.e., the models from a given class cluster better) when $ \mathbf{X} $ contains prompts relevant to the class-wise differences.
More details related to the models shown in the two perspective spaces are provided in Appendix \ref{app:fine-tuning}.

The perspective space that includes the entire history of a system can be learned by considering the CMDS of the  $ |\mathcal{F}|^{T}~\times~|\mathcal{F}|^{T}$ pairwise distance matrix with entries $ ||\tilde{g}(f_{i}^{(t)}(\mathbf{X})) - \tilde{g}(f_{j}^{(t')}(\mathbf{X}))||_{F} $ for all $ i, j~\in~\{1, \hdots, |\mathcal{F}|\} $ and all $ t, t'~\in~\{1, \hdots, T\} $. We use this perspective space when studying the systems below. The methodology can be extended to instances where only a partial history of the system is observed via out-of-sample methods \citep{bengio2003out, levin2018out}.

\section{Simulating systems of interacting LLMs}
\label{sec:simulations}

We next simulate three different systems of interacting LLMs to demonstrate the effectiveness of the perspective space and its derivatives for capturing model and system dynamics for different underlying communication structures.
The initial models in each system are based on an instance of the 410-million parameter model from the Pythia suite \citep{biderman2023pythia} that has been instruction-tuned using Databricks' Dolly 15k \citep{DatabricksBlog2023DollyV2}. 
For each system we further fine-tune the base model on random question-pairs from setting specific topics from Yahoo! Answers (YA) dataset \citep{zhang2015character} to promote response variation. 
We provide details on the instruction-tuning of the base model and the fine-tuning of the initial models in Appendix \ref{app:insruction-tuning} and Appendix \ref{app:fine-tuning}, respectively. We use \texttt{all-MiniLM-L6-v2}, a sentence embedding function from \citep{reimers2019sentence} based on \citep{wang2020minilm} hosted on the HuggingFace Hub \citep{wolf2020huggingfaces}, as the surrogate embedding function and the implementation of CMDS from Graspologic \citep{chung2019graspy}.

\begin{figure*}[t!]
\centering
\captionsetup[subfigure]{justification=centering}
\begin{subfigure}{0.49\textwidth}
    \centering
\includegraphics[width=\linewidth]{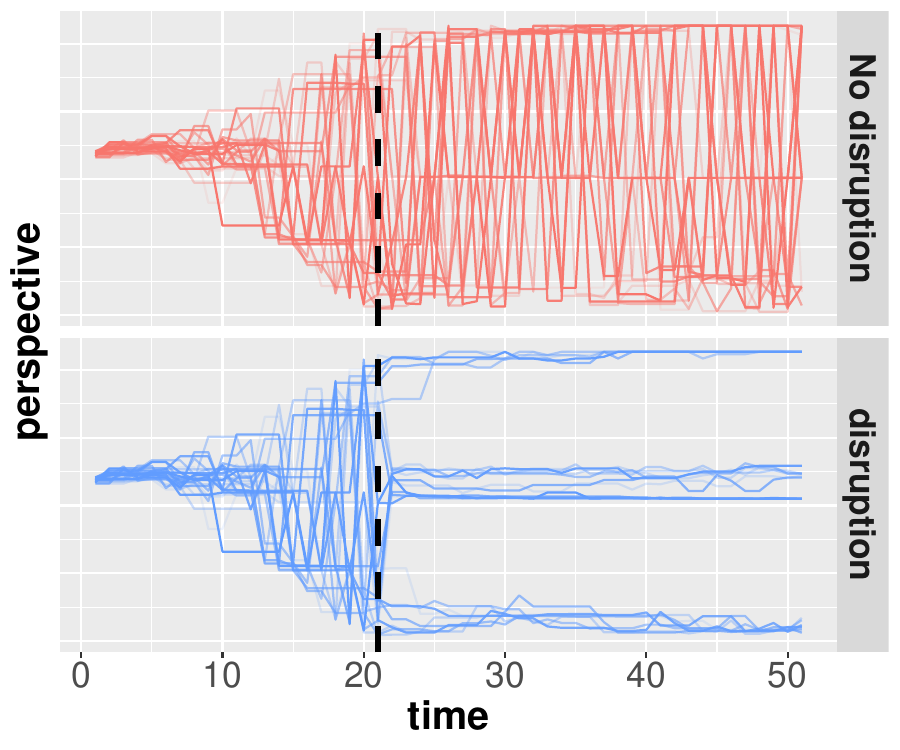}
\end{subfigure}
\begin{subfigure}{0.49\textwidth}
    \centering
    \includegraphics[width=\linewidth]{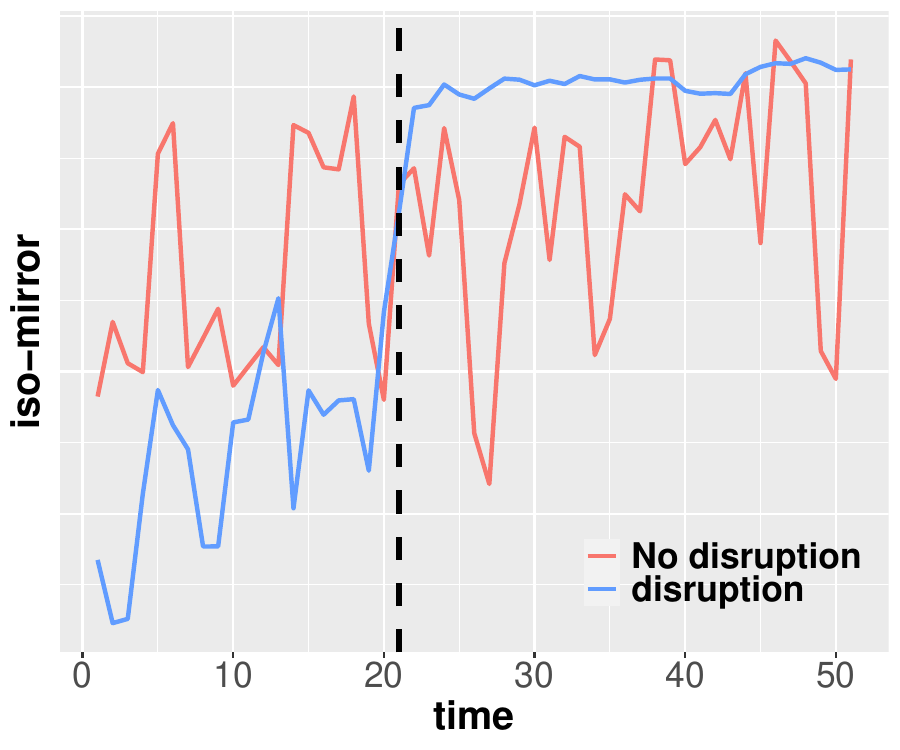}
    \end{subfigure}
\caption{Tracking individual perspective (left) and system-level dynamics (right) of communication networks of chat-based language models with (bottom left) and without (top left) a disruption in communication structure.}
\label{fig:change-point}
\end{figure*}

In the three Case Studies (C.S.) we consider, each model interacts with another model in the system at each $ t $. 
An interaction consists of model $ i $ asking model $ j \neq i $ a random set of questions from a fixed question bank and fine-tuning model $ i $ using the resulting question-answer pairs as fine-tuning data.
For a given $ t $, the underlying communication structure $ E^{(t)} $ determines which set of model interactions are possible for model $ i $. In particular, the interviewed model $ j $ is randomly selected from the set of models such that $ (f_{j}, f_{i}) \in E^{(t)} $. The fixed question bank is used as the evaluation set to induce the perspective space. 

While each system that we study technically consists of models and databases, each dataset is associated with only a single model. For convenience we discuss the systems as if the models themselves are directly connected.
Our setting -- where models are sequentially trained on each others outputs without intervention -- can be viewed as a generalization of a single model sequentially trained on its own outputs as studied in \citep{shumailov2024curse}.

At the end of each simulation setting we provide examples that motivated the case study.


\subsubsection*{C.S. 1: Disrupting the communication network}
We first study a system with $ |\mathcal{F}| = 25 $ 
models fine-tuned on different 400 random samples from YA 
with topic ``Society \& Culture" under two different system evolutions.
For the first system evolution
\begin{wrapfigure}{r}{0.5\textwidth} 
\begin{center}
\includegraphics[width=0.5\textwidth]{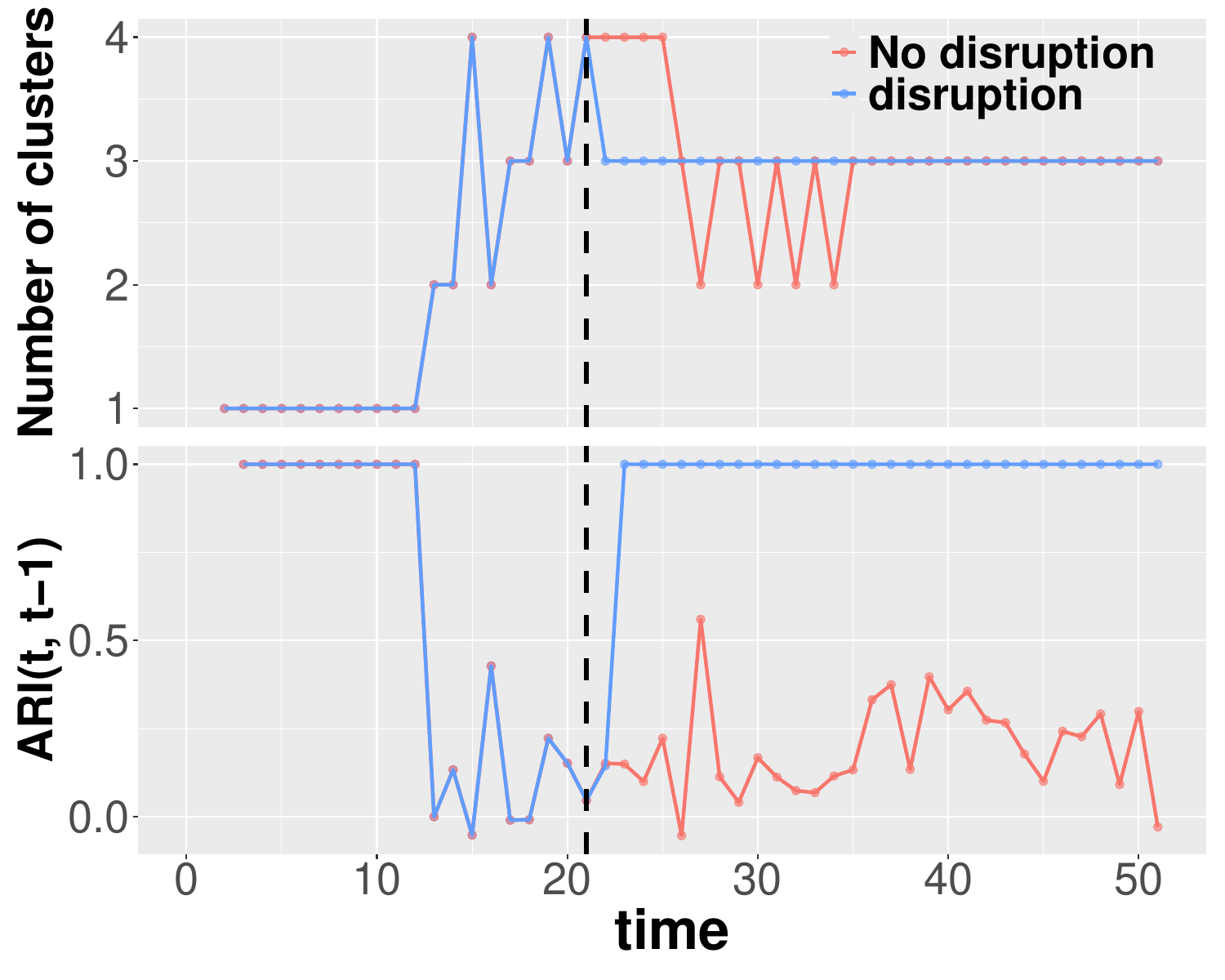}
  \end{center}
  \caption{Estimated number of clusters found via GMM with BIC (top) and sequential ARI of cluster labels (bottom) for disrupted and undisrupted systems. The number of clusters in both systems stabilize, indicating the presence of model sinks. Model sinks are unstable in a system with no disruption and stable in a system with a disruption.}
  \label{fig:clusters-ari}
\end{wrapfigure}
the underlying communication structure is unrestricted (i.e., $ E^{(t)} $ fully connected, see Figure \ref{fig:network-examples} ``fully connected") for all $ t $. For the second system evolution the underlying communication structure is unrestricted for $ t < t^{*} $ and is then local-only (i.e., $ (f_{i}, f_{j}) \in E^{(t)} $ only if model $ i $ is model $ j $'s nearest neighbor in perspective space after the interactions at $ t-1 $) thereafter.
We refer to the shift from unrestricted communication to local communication as a disruption in the communication structure. 

At each time $ t $ model $ i $ 
asks 50 random questions from a question bank of 400 questions from YA with topic ``Society \& Culture". 
The initial 1-d perspectives of the models are relatively close to each other, as can be seen at $ t = 0 $ in both the top left and bottom left figures of Figure \ref{fig:change-point}.
As the system evolves for $ t < t^{*} $, we observe the models ``exploring" the perspective space.
For the system that does not experience a disruption (top left), the exploration in perspective eventually stagnates and each model appears to oscillate between three different global perspective ``sinks", one near the top of the figure, one in the middle of the figure, and one near the bottom of the figure. 
For the system that experiences a disruption at $ t^{*} = 21 $ (bottom left) the exploration in perspective space similarly stops, though the models do not oscillate between global sinks and, instead, persist in local sinks.
The existence of multiple model sinks in both evolutions generalizes the behavior observed in \citep{shumailov2024curse}, where the sequence of a single model sequentially trained on its own output converges to a degenerate model sink.

The difference in local and global sinks is quantified in Figure \ref{fig:clusters-ari}, where we report the number of clusters at each $ t $ and the similarity of sequential cluster labels. 
We use Gaussian Mixture Modeling with the Bayesian Information Criterion (BIC) to estimate the number of clusters \citep{doi:10.1198/016214502760047131} and adjusted Rand index (ARI) to measure cluster label similarity.
We find that the number of clusters for both systems eventually stabilizes and that the ARI between 
sequential cluster labels is lower for the global communication network after stabilization, which signifies higher cluster instability.


We quantify the general evolution of the systems via the ``iso-mirror" \citep{athreya2022discovering} in the right figure of Figure \ref{fig:change-point}. 
The iso-mirror is a system-level summary of the dynamics that takes into account the collection of model-level dynamics. 
In our setting, the iso-mirror corresponding to the system that does not experience a disruption is unstable throughout $ t $.
The iso-mirror corresponding to the disrupted system, however, clearly changes behavior at $ t^{*} $ and remains constant throughout the remainder of its evolution.

\noindent \textbf{Motivating examples.} This case study was largely motivated by the COVID-19 pandemic \citep{zuzul2023dynamic} where social distancing, work from home, and social pods changed the latent communication structure for entire communities.
It is also relevant to communication networks for range-limited devices where the definition of ``local" depends on the geographical location of the device \citep{9044329}.

\subsubsection*{C.S. 2: Diffusion of an adversarial perspective}

\begin{figure*}[h!]
\centering
\captionsetup[subfigure]{justification=centering}
\begin{subfigure}{0.9\textwidth}
    \centering
    \includegraphics[width=\linewidth]{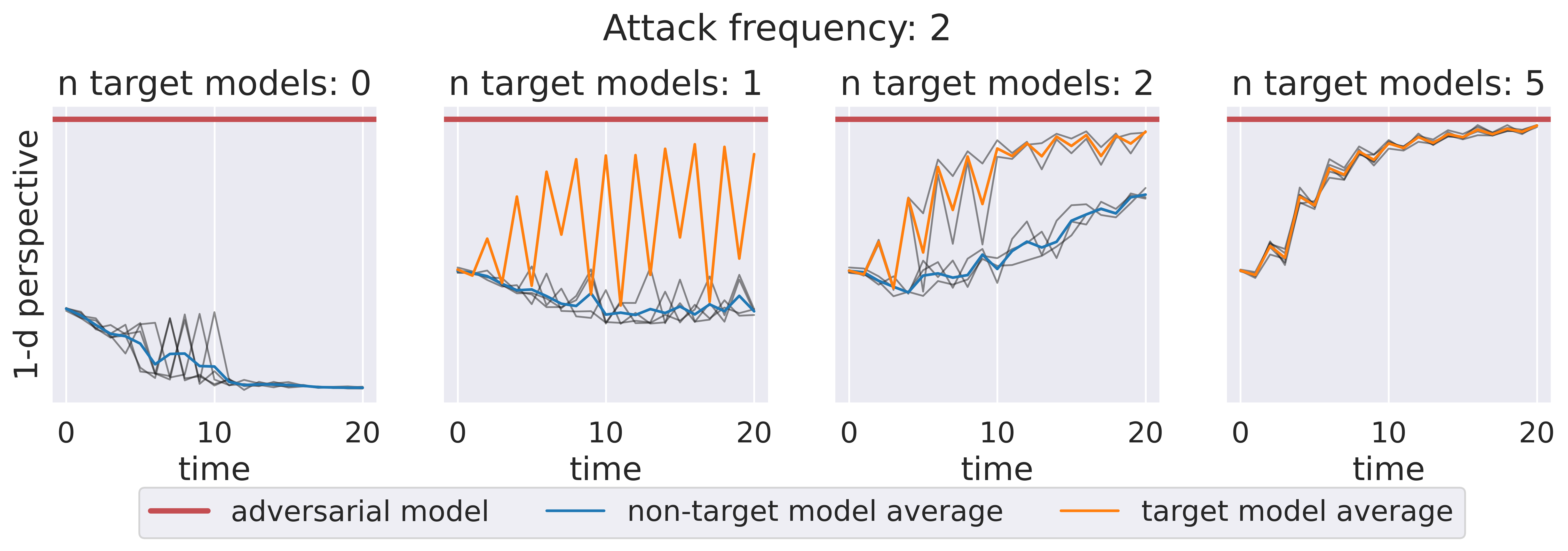}
\end{subfigure} \\
\begin{subfigure}{0.9\textwidth}
    \centering
    \includegraphics[width=\linewidth]{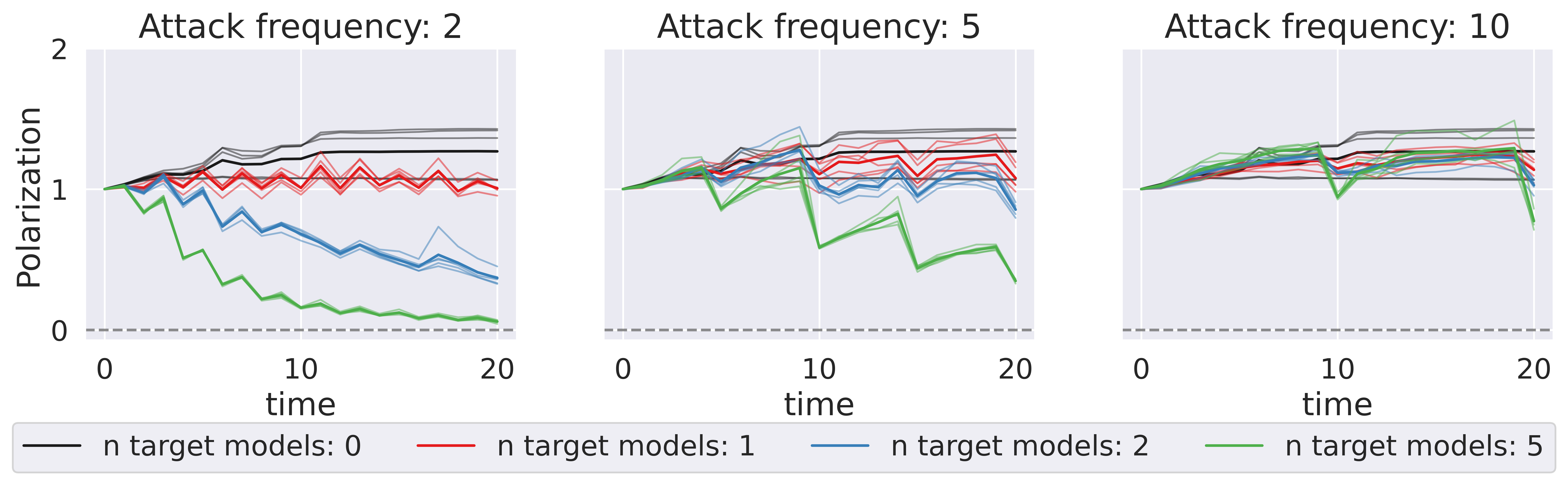}
    \end{subfigure}
\caption{
The evolution of 1-d perspectives of five interacting models where two models interact with an ``adversarial" model every other interaction (top). Given enough nodes to influence, the adversarial model can compromise the entire network -- as captured by the difference between the average 1-d perspective of the non-adversarial models and the 1-d perspective of the adversarial model for various amounts of target models and various attack frequencies (bottom).}
\label{fig:adversarial-experiment}
\end{figure*}

We next consider a system with $| \mathcal{F} | = 6 $ models where five of the models are fine-tuned on a random set of 1000 question-answer pairs from YA with topic ``Society \& Culture" and the sixth is fine-tuned on a random set of 1000 question-answer pairs from YA with topic ``Science \& Mathematics". 
We refer to the model trained on data with topic ``Science \& Mathematics" as an ``adversarial" model since it does not share the same initial perspective as the other five in expectation. 
A non-adversarial model is referred to as a ``target" model at time $ t $ if there is an edge from the adversarial model to it in $ E^{(t)} $. 
Target models are randomly selected at the beginning of the evolution of the system and remain targets throughout a simulation. 
The evaluation set consists of 200 questions from the ``Science \& Mathematics" topic. 
At each iteration model $ i $ asks model $ j $ 100 questions.

For this experiment $ E^{(t)} $ oscillates between two states. 
The first is a base state where the non-adversarial subnetwork is fully connected and there are no edges to or from the adversarial model. 
The second is a ``vulnerable" state where there is an edge from the adversarial model to all target models, there are no other in-bound edges to the adversarial or target models, the non-target non-adversarial subnetwork is fully connected, and there are edges from the target models to the non-target models (see Figure \ref{fig:network-examples} ``vulnerable"). 
We simulate systems that have a vulnerable communication network once every two, five or ten iterations.

The trajectories of the 1-d perspectives of the models in the system with a vulnerable communication every other iteration are shown in the top of Figure \ref{fig:adversarial-experiment} for systems with 0, 1, 2 and 5 targets. 
We also report the average perspective of the targeted models and the average perspective of the non-targeted models for each system.  

For the system with no targets (top left) we observe similar behavior to the first case study under no disruption: the models initially explore the perspective space and eventually settle in a model sink. 
For the system with a single target we see the targeted model (top center left) oscillate between the adversarial perspective and the average perspective of the non-targeted models. 
Non-target models that interact with the target models immediately after the communication network was vulnerable are similarly pulled towards the adversarial perspective but to a lesser extent. 
Together these two effects limit the perspective exploration of the models in the system and eliminate the presence of the model sink.

For the system with two targets (top center right) the targeted models oscillate between the adversarial perspective and the average non-target perspective but the oscillations dampen as the non-target model pespectives start to drift towards the adversarial perspective. By $ t = 20 $ the average non-target perspective is closer to the adversarial perspective than its own starting position. That is, the entire system of LLMs has been compromised by the adversarial model targeting only a \textit{minority} of the models in the system. The average perspective of models in a system with five targets (top right) quickly approaches the adversarial perspective.

In this setting we summarize system behavior via polarization defined as the difference in the average perspective of non-adversarial models and the perspective of the adversarial model normalized by this difference at $ t = 0 $. 
We report the polarization for five initialization for vulnerable communication frequencies of two, five, and ten in the bottom of Figure \ref{fig:adversarial-experiment}. 
For example, for an attack frequency of two we see that polarization neatly summarizes our observations. In particular, the polarization increases when there are no target models, the polarization is relatively stable when there is a single target, the polarization slowly drifts towards zero when there are two targets, and the polarization quickly approaches zero when there are five targets. The system is more susceptible when more models are targeted for attack frequencies of five and ten, as well.

The trend across attack frequencies for a fixed number of target models indicates that given enough time between attacks the average model perspective is able to recover. This is likely due to the interaction mechanic involving a random subset of the evaluation questions -- instead of the entire set -- that enables system-level perspective homeostasis.

\noindent \textbf{Motivating example.} This case study was designed to mimic information diffusion in the presence of simple propaganda machines and to study how ``attacks" on a minority affects the entire system. 

\subsubsection*{C.S. 3: Mitigating or promoting polarization}

\begin{figure*}[h]
\centering
\captionsetup[subfigure]{justification=centering}
\begin{subfigure}{0.49\textwidth}
    \centering
    \includegraphics[width=\linewidth]{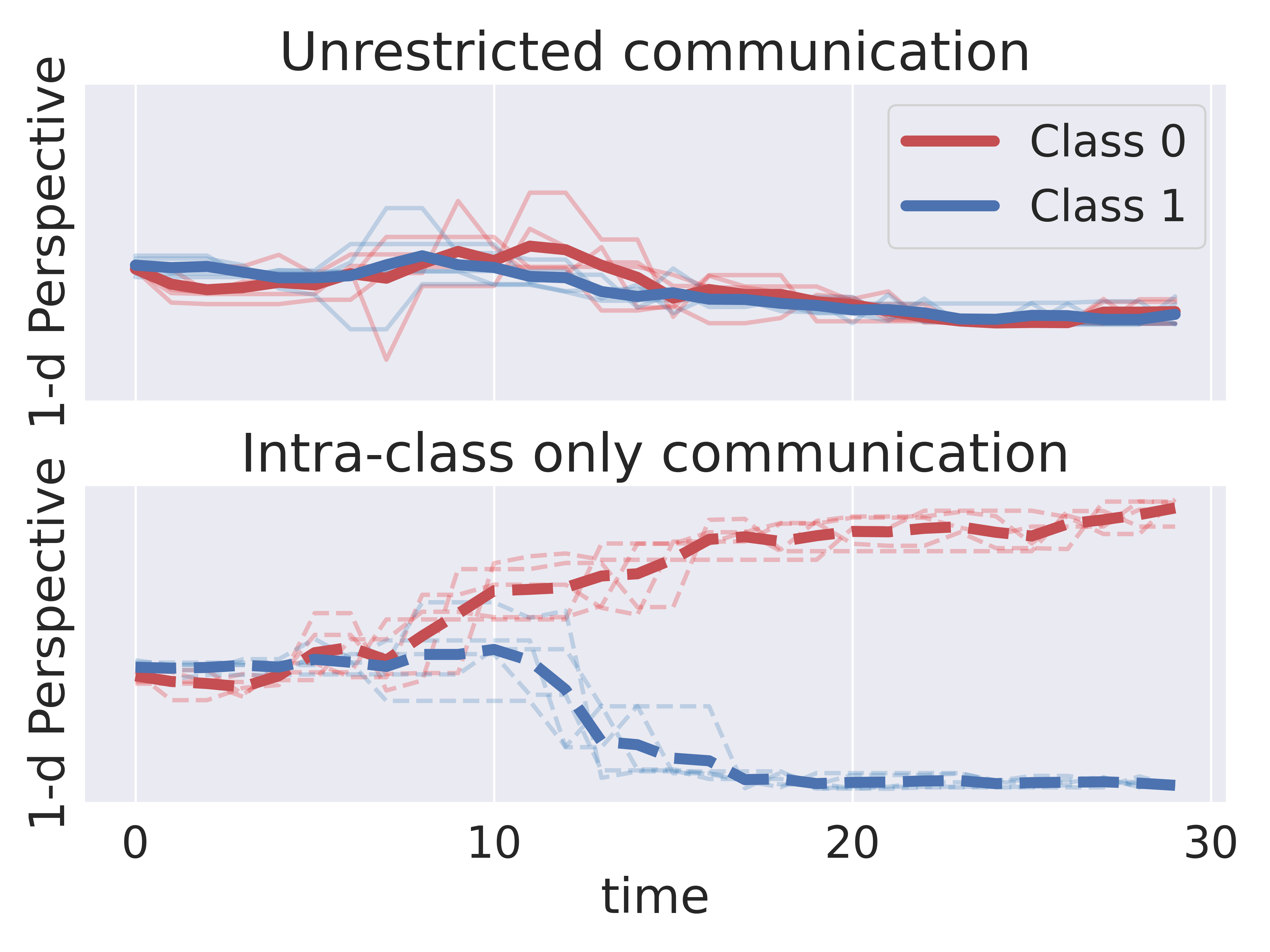}
\end{subfigure}
\begin{subfigure}{0.49\textwidth}
    \centering
    \includegraphics[width=\linewidth]{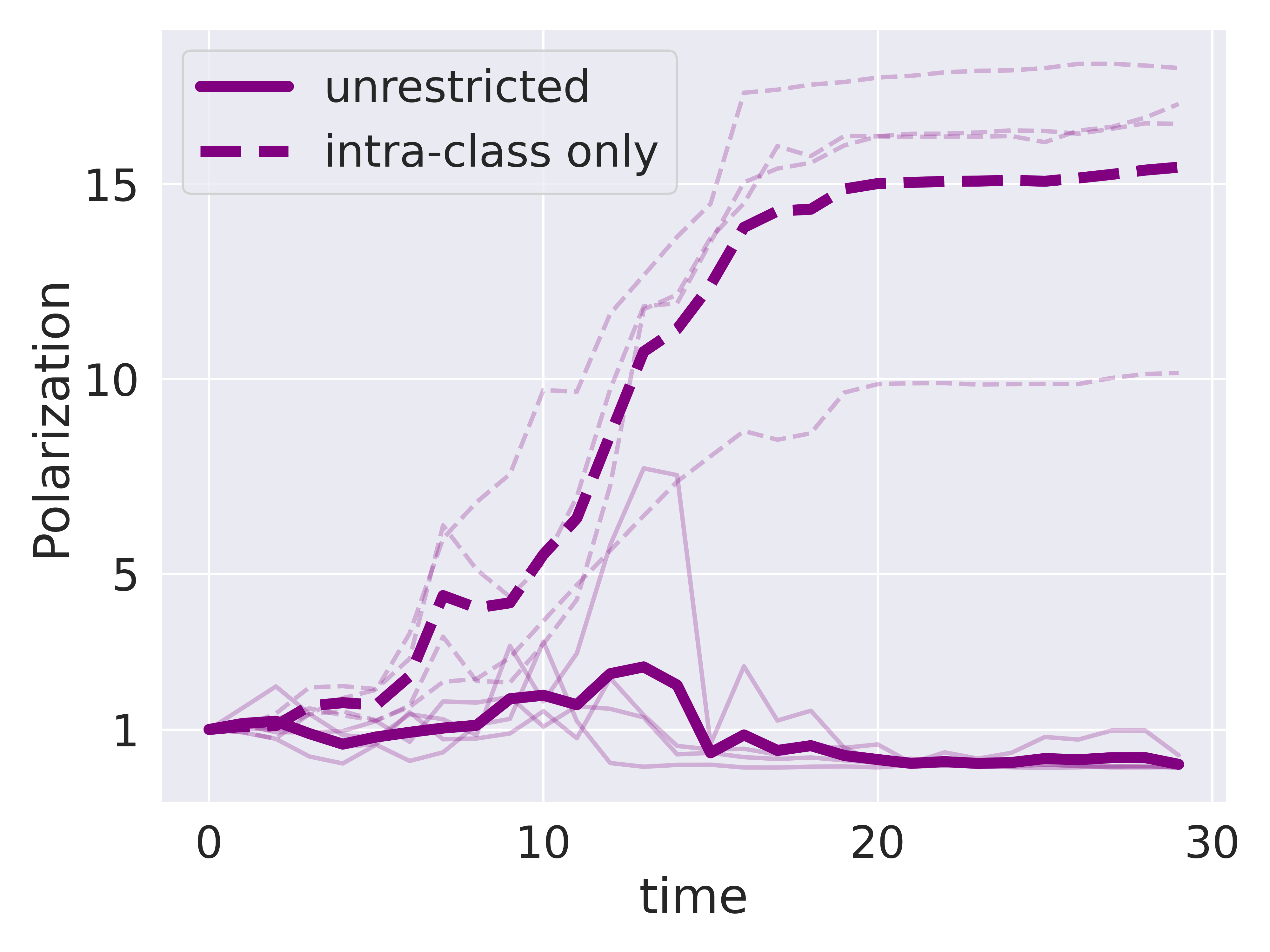}
    \end{subfigure}
\caption{
The evolution of 1-d perspective space representations of ten models from two classes under different underlying communication structures -- unrestricted (left, top) and intra-class only (left, bottom). 
Class-wise average 1-d perspectives (bolded) are intertwined throughout the evolution of the system with unrestricted communication and diverge with intra-class only communication.
Polarization captures this difference in behavior over multiple iterations of the experiment (right).}
\label{fig:echo-chambers}
\end{figure*}

In our last case study we consider a system of $ | \mathcal{F} | = 10 $ models where five of the models are fine-tuned on 1000 random question-answer pairs from YA with topic ``Society \& Culture" and the other five are fine-tuned on 1000 random question-answer pairs from YA with topic ``Science \& Mathematics" . 
We let the topic in which the fine-tuning data is sampled from parameterize model ``class". 
The evaluation set consists of 200 questions from each class. An interaction consists of model $ i $ asking model $ j $ 100 questions.

In this experiment we consider two different communication structures: unrestricted communication where $ E^{(t)} $ is fully connected and intra-class only communication where $ E^{(t)} $ consists of two unconnected class-wise fully connected subnetworks (see Figure \ref{fig:network-examples} ``intra-class only").
A system has the same communication structure for the entirety of its evolution.
The top left figure of Figure \ref{fig:echo-chambers} shows 1-d perspectives of the models in the system with unrestricted communication. Bolded lines represent the class average.
As with fully connected communication network settings in the other case studies, we observe a period of perspective exploration before stabilizing. Notably, the two class-means stay intertwined throughout the entirety of the evolution of the system.

The bottom left figure of Figure \ref{fig:echo-chambers} shows the evolution of 1-d perspectives with intra-class only communication. Under the intra-class only regime we see that the two classes explore \textit{different} regions of the perspective space and eventually settle into two sinks with a much greater distance between them then the class-wise differences at $ t = 0 $. The polarization of the class-wise averages captures the distancing of the perspective ``echo chambers", as reported in the right figure of Figure \ref{fig:echo-chambers}. Indeed, the polarization increased by 15x on average over four different simulation initializations under intra-class only communication. 
Average polarization is near zero by the end of the simulations under unrestricted communication.

\noindent \textbf{Motivating example.} This case study was designed to investigate the effect of extreme underlying communication networks on two party systems.

\section{Related Work}

Our work is closely related to simulating groups of computational agents to study sociological and cultural phenomena  \citep{steels1990cooperation, wagner2003progress} and to continual learning \citep{vogelstein2020general, geisa2021towards}. The former has seen renewed interest with the recent successes of LLMs. 
In particular, LLMs are -- as of this writing -- the computational tool that produces language artifacts most similar to ours and, as such, are an intriguing prospect for multi-agent sociological and cultural simulations. 
Recent work has included objective-less behavioral studies \citep{park2023generative}, studying the formation of social networks \citep{papachristou2024network}, tracking opinion dynamics via classification of LLM response \citep{chuang2023simulating}, and analyzing document collaboration \citep{perez2024cultural}.
Our work extends these by introducing a framework to systematically study interventions and by introducing a quantitative method for tracking the evolution of agent perspectives.

Continual learning \citep{thrun1995learning, thrun1998lifelong} is largely concerned with how a single agent adapts to previously unseen inference tasks while avoiding ``catastrophically forgetting" \citep{mccloskey1989catastrophic, kirkpatrick2017overcoming} previous tasks. 
Our setting is slightly different, since we have multiple agents and no explicit task -- though a large movement in perspective space is likely highly correlated to change in performance on language benchmarks related to the evaluation set.
Indeed, large enough movements in perspective space and the emergence of model sinks when training a model recursively is related to catastrophic forgetting \citep{shumailov2024curse}.

\section{Conclusion}
\label{sec:conclusion}

We introduced a system-of-LLMs-as-a-graph to enable systematic interventions to a system of interacting LLMs and the perspective space to quantitatively study the corresponding evolution of the system.
We used these tools to highlight differences in paired systems across three case studies. 
For the particular interaction mechanic and update function that we used in our simulations, the model behaviors in perspective space consistently demonstrated initial model exploration and, in most cases, the emergence and persistence of model sinks.
Further, we used derivatives of the perspective space such as the iso-mirror, polarization, and clustering to highlight differences in the evolution of paired systems. 

For example, we observed differences in the iso-mirror (stable versus unstable after disruption) and clustering (global sinks versus local sinks after disruption) in the first case study; differences in the sensitivity of the average perspective of non-adversarial models to an adversarial perspective across number of victims and frequency of attack in the second case study; and differences in the behavior of polarization of two classes of models in the third case study.

\section{Limitations}
\label{sec:limitations}
A system of interacting language models is a complicated system and, as such, analysis of them will often require simplification of aspects of the system. 
Our case studies are no expection.
For example, the interaction mechanic (i.e., each model interacts with exactly one of its neighbors at time $ t $) and update function (i.e., update model weights via fine-tuning) used in the simulations are more proof-of-concept than final-product in that they do not reflect our beliefs on how individuals within a community interact or ``update" themselves, nor are currently deployed models constantly updated. 
While we do not attempt to enumerate all possible improvements here, we believe that it is imperative to work closely with social and cognitive scientists to understand the appropriateness of considering systems of LLMs as a proxy for human communities or online forums before generalizing observed simulated behavior to human-facing communities. 
Future work along these lines will include two major fronts: i) designing comprehensive statistical frameworks to understand the appropriateness of using a system of interacting LLMs as a proxy for various social settings and ii) extending simulation settings to include more sociologically plausible interaction and update mechanics.

Further, the simulation studies herein are but three system configurations worth considering.  
Indeed, of immediate interest is an extension to hierarchical social structures observed in large commercial and government institutions where the perspective space can be used to understand the effect of information injection, re-organizations, third-party seminars, etc. on individual-level, team-level, and organization-level dynamics.

There are also limitations related to the analysis of each of the three case studies we presented.
For example, for the first case study we only investigated the difference between system behavior of global communication and global to hyper-local communication. 
More nuanced investigations into the effect of the number of models, the effect of the initializations of the models, the effect of the definition of ``local", etc. is necessary to understand how the empirical observations may generalize to the real world.
Similarly, for the second case study we only considered a single static adversarial model.
A more realistic simulation may include multiple dynamic adversarial models.
For the third case study, if this analysis is to be used to understand polarization of political parties, it is necessary to understand the effect of cross-party communication, however rare it may be.
We, again, believe that it is necessary to comprehensively explore each of these experiments before making claims about its applicability to society and human-model forums.

Lastly, we introduce the perspective space and demonstrate that it is sensitive to evaluation set. 
We do not, however, comprehensively explore or discuss potential applications or alternative model-based similarities.
Similar methods have been used 
We expect the perspective space to be useful for various model-level inference tasks, as similar methods have been successfully used for classification \citep{chen2022mental} and change-point detection \citep{chen2023discovering} in neuroscience applications.
We also expect the model-based similarity most effective for capturing model differences will be system and task dependent \citep{eaton2008modeling, zamir2018taskonomy, helm2020partition}.





\label{method}

\noindent \textbf{Acknowledgements.} We would like to thank Avanti Athreya, Henry Farrell, Hahrie Han, Teresa Huang, Vince Lyzinski, Harvey McGuinness, and Tim Wang for their helpful feedback and discussions throughout the development of this manuscript.

\bibliographystyle{iclr2021/iclr2021_conference}
\bibliography{biblio}

\begin{thebibliography}{44}
\providecommand{\natexlab}[1]{#1}
\providecommand{\url}[1]{\texttt{#1}}
\expandafter\ifx\csname urlstyle\endcsname\relax
  \providecommand{\doi}[1]{doi: #1}\else
  \providecommand{\doi}{doi: \begingroup \urlstyle{rm}\Url}\fi

\bibitem[Achiam et~al.(2023)Achiam, Adler, Agarwal, Ahmad, Akkaya, Aleman, Almeida, Altenschmidt, Altman, Anadkat, et~al.]{achiam2023gpt}
Josh Achiam, Steven Adler, Sandhini Agarwal, Lama Ahmad, Ilge Akkaya, Florencia~Leoni Aleman, Diogo Almeida, Janko Altenschmidt, Sam Altman, Shyamal Anadkat, et~al.
\newblock Gpt-4 technical report.
\newblock \emph{arXiv preprint arXiv:2303.08774}, 2023.

\bibitem[Athreya et~al.(2022)Athreya, Lubberts, Park, and Priebe]{athreya2022discovering}
Avanti Athreya, Zachary Lubberts, Youngser Park, and Carey~E Priebe.
\newblock Discovering underlying dynamics in time series of networks.
\newblock \emph{arXiv preprint arXiv:2205.06877}, 2022.

\bibitem[Bengio et~al.(2003)Bengio, Paiement, Vincent, Delalleau, Roux, and Ouimet]{bengio2003out}
Yoshua Bengio, Jean-fran{\c{c}}cois Paiement, Pascal Vincent, Olivier Delalleau, Nicolas Roux, and Marie Ouimet.
\newblock Out-of-sample extensions for lle, isomap, mds, eigenmaps, and spectral clustering.
\newblock \emph{Advances in neural information processing systems}, 16, 2003.

\bibitem[Biderman et~al.(2023)Biderman, Schoelkopf, Anthony, Bradley, O’Brien, Hallahan, Khan, Purohit, Prashanth, Raff, et~al.]{biderman2023pythia}
Stella Biderman, Hailey Schoelkopf, Quentin~Gregory Anthony, Herbie Bradley, Kyle O’Brien, Eric Hallahan, Mohammad~Aflah Khan, Shivanshu Purohit, USVSN~Sai Prashanth, Edward Raff, et~al.
\newblock Pythia: A suite for analyzing large language models across training and scaling.
\newblock In \emph{International Conference on Machine Learning}, pp.\  2397--2430. PMLR, 2023.

\bibitem[Brinkmann et~al.(2023)Brinkmann, Baumann, Bonnefon, Derex, Müller, Nussberger, Czaplicka, Acerbi, Griffiths, Henrich, Leibo, McElreath, Oudeyer, Stray, and Rahwan]{Brinkmann_2023}
Levin Brinkmann, Fabian Baumann, Jean-François Bonnefon, Maxime Derex, Thomas~F. Müller, Anne-Marie Nussberger, Agnieszka Czaplicka, Alberto Acerbi, Thomas~L. Griffiths, Joseph Henrich, Joel~Z. Leibo, Richard McElreath, Pierre-Yves Oudeyer, Jonathan Stray, and Iyad Rahwan.
\newblock Machine culture.
\newblock \emph{Nature Human Behaviour}, 7\penalty0 (11):\penalty0 1855–1868, November 2023.
\newblock ISSN 2397-3374.
\newblock \doi{10.1038/s41562-023-01742-2}.
\newblock URL \url{http://dx.doi.org/10.1038/s41562-023-01742-2}.

\bibitem[Chen et~al.(2022)Chen, Helm, Lytvynets, Yang, and Priebe]{chen2022mental}
Guodong Chen, Hayden~S Helm, Kate Lytvynets, Weiwei Yang, and Carey~E Priebe.
\newblock Mental state classification using multi-graph features.
\newblock \emph{Frontiers in Human Neuroscience}, 16:\penalty0 930291, 2022.

\bibitem[Chen et~al.(2023)Chen, Park, Saad-Eldin, Lubberts, Athreya, Pedigo, Vogelstein, Puppo, Silva, Muotri, et~al.]{chen2023discovering}
Tianyi Chen, Youngser Park, Ali Saad-Eldin, Zachary Lubberts, Avanti Athreya, Benjamin~D Pedigo, Joshua~T Vogelstein, Francesca Puppo, Gabriel~A Silva, Alysson~R Muotri, et~al.
\newblock Discovering a change point in a time series of organoid networks via the iso-mirror.
\newblock \emph{arXiv preprint arXiv:2303.04871}, 2023.

\bibitem[Chuang et~al.(2023)Chuang, Goyal, Harlalka, Suresh, Hawkins, Yang, Shah, Hu, and Rogers]{chuang2023simulating}
Yun-Shiuan Chuang, Agam Goyal, Nikunj Harlalka, Siddharth Suresh, Robert Hawkins, Sijia Yang, Dhavan Shah, Junjie Hu, and Timothy~T Rogers.
\newblock Simulating opinion dynamics with networks of llm-based agents.
\newblock \emph{arXiv preprint arXiv:2311.09618}, 2023.

\bibitem[Chung et~al.(2019)Chung, Pedigo, Bridgeford, Varjavand, Helm, and Vogelstein]{chung2019graspy}
Jaewon Chung, Benjamin~D Pedigo, Eric~W Bridgeford, Bijan~K Varjavand, Hayden~S Helm, and Joshua~T Vogelstein.
\newblock Graspy: Graph statistics in python.
\newblock \emph{Journal of Machine Learning Research}, 20\penalty0 (158):\penalty0 1--7, 2019.

\bibitem[Conover et~al.(2023)Conover, Hayes, Mathur, Xie, Wan, Shah, Ghodsi, Wendell, Zaharia, and Xin]{DatabricksBlog2023DollyV2}
Mike Conover, Matt Hayes, Ankit Mathur, Jianwei Xie, Jun Wan, Sam Shah, Ali Ghodsi, Patrick Wendell, Matei Zaharia, and Reynold Xin.
\newblock Free dolly: Introducing the world's first truly open instruction-tuned llm, 2023.
\newblock URL \url{https://www.databricks.com/blog/2023/04/12/dolly-first-open-commercially-viable-instruction-tuned-llm}.

\bibitem[Devlin et~al.(2018)Devlin, Chang, Lee, and Toutanova]{devlin2018bert}
Jacob Devlin, Ming-Wei Chang, Kenton Lee, and Kristina Toutanova.
\newblock Bert: Pre-training of deep bidirectional transformers for language understanding.
\newblock \emph{arXiv preprint arXiv:1810.04805}, 2018.

\bibitem[Duderstadt et~al.(2023)Duderstadt, Helm, and Priebe]{duderstadt2023comparing}
Brandon Duderstadt, Hayden~S Helm, and Carey~E Priebe.
\newblock Comparing foundation models using data kernels.
\newblock \emph{arXiv preprint arXiv:2305.05126}, 2023.

\bibitem[Eaton et~al.(2008)Eaton, Desjardins, and Lane]{eaton2008modeling}
Eric Eaton, Marie Desjardins, and Terran Lane.
\newblock Modeling transfer relationships between learning tasks for improved inductive transfer.
\newblock In \emph{Machine Learning and Knowledge Discovery in Databases: European Conference, ECML PKDD 2008, Antwerp, Belgium, September 15-19, 2008, Proceedings, Part I 19}, pp.\  317--332. Springer, 2008.

\bibitem[Fraley \& Raftery(2002)Fraley and Raftery]{doi:10.1198/016214502760047131}
Chris Fraley and Adrian~E Raftery.
\newblock Model-based clustering, discriminant analysis, and density estimation.
\newblock \emph{Journal of the American Statistical Association}, 97\penalty0 (458):\penalty0 611--631, 2002.
\newblock \doi{10.1198/016214502760047131}.
\newblock URL \url{https://doi.org/10.1198/016214502760047131}.

\bibitem[Geisa et~al.(2021)Geisa, Mehta, Helm, Dey, Eaton, Dick, Priebe, and Vogelstein]{geisa2021towards}
Ali Geisa, Ronak Mehta, Hayden~S Helm, Jayanta Dey, Eric Eaton, Jeffery Dick, Carey~E Priebe, and Joshua~T Vogelstein.
\newblock Towards a theory of out-of-distribution learning.
\newblock \emph{arXiv preprint arXiv:2109.14501}, 2021.

\bibitem[Helm et~al.(2023)Helm, Priebe, and Yang]{helm2023statistical}
Hayden Helm, Carey~E Priebe, and Weiwei Yang.
\newblock A statistical turing test for generative models.
\newblock \emph{arXiv preprint arXiv:2309.08913}, 2023.

\bibitem[Helm et~al.(2020)Helm, Mehta, Duderstadt, Yang, White, Geisa, Vogelstein, and Priebe]{helm2020partition}
Hayden~S Helm, Ronak~D Mehta, Brandon Duderstadt, Weiwei Yang, Christoper~M White, Ali Geisa, Joshua~T Vogelstein, and Carey~E Priebe.
\newblock A partition-based similarity for classification distributions.
\newblock \emph{arXiv preprint arXiv:2011.06557}, 2020.

\bibitem[Jumper et~al.(2021)Jumper, Evans, Pritzel, Green, Figurnov, Ronneberger, Tunyasuvunakool, Bates, {\v{Z}}{\'\i}dek, Potapenko, et~al.]{jumper2021highly}
John Jumper, Richard Evans, Alexander Pritzel, Tim Green, Michael Figurnov, Olaf Ronneberger, Kathryn Tunyasuvunakool, Russ Bates, Augustin {\v{Z}}{\'\i}dek, Anna Potapenko, et~al.
\newblock Highly accurate protein structure prediction with alphafold.
\newblock \emph{Nature}, 596\penalty0 (7873):\penalty0 583--589, 2021.

\bibitem[Kirkpatrick et~al.(2017)Kirkpatrick, Pascanu, Rabinowitz, Veness, Desjardins, Rusu, Milan, Quan, Ramalho, Grabska-Barwinska, et~al.]{kirkpatrick2017overcoming}
James Kirkpatrick, Razvan Pascanu, Neil Rabinowitz, Joel Veness, Guillaume Desjardins, Andrei~A Rusu, Kieran Milan, John Quan, Tiago Ramalho, Agnieszka Grabska-Barwinska, et~al.
\newblock Overcoming catastrophic forgetting in neural networks.
\newblock \emph{Proceedings of the national academy of sciences}, 114\penalty0 (13):\penalty0 3521--3526, 2017.

\bibitem[Kosinski(2023)]{kosinski2023evaluating}
Michal Kosinski.
\newblock Evaluating large language models in theory of mind tasks.
\newblock \emph{arXiv e-prints}, pp.\  arXiv--2302, 2023.

\bibitem[Levin et~al.(2018)Levin, Roosta, Mahoney, and Priebe]{levin2018out}
Keith Levin, Fred Roosta, Michael Mahoney, and Carey Priebe.
\newblock Out-of-sample extension of graph adjacency spectral embedding.
\newblock In \emph{International Conference on Machine Learning}, pp.\  2975--2984. PMLR, 2018.

\bibitem[McCloskey \& Cohen(1989)McCloskey and Cohen]{mccloskey1989catastrophic}
Michael McCloskey and Neal~J Cohen.
\newblock Catastrophic interference in connectionist networks: The sequential learning problem.
\newblock In \emph{Psychology of learning and motivation}, volume~24, pp.\  109--165. Elsevier, 1989.

\bibitem[Nussbaum et~al.(2024)Nussbaum, Morris, Duderstadt, and Mulyar]{nussbaum2024nomic}
Zach Nussbaum, John~X. Morris, Brandon Duderstadt, and Andriy Mulyar.
\newblock Nomic embed: Training a reproducible long context text embedder, 2024.

\bibitem[Oquab et~al.(2023)Oquab, Darcet, Moutakanni, Vo, Szafraniec, Khalidov, Fernandez, Haziza, Massa, El-Nouby, et~al.]{oquab2023dinov2}
Maxime Oquab, Timoth{\'e}e Darcet, Th{\'e}o Moutakanni, Huy Vo, Marc Szafraniec, Vasil Khalidov, Pierre Fernandez, Daniel Haziza, Francisco Massa, Alaaeldin El-Nouby, et~al.
\newblock Dinov2: Learning robust visual features without supervision.
\newblock \emph{arXiv preprint arXiv:2304.07193}, 2023.

\bibitem[Papachristou \& Yuan(2024)Papachristou and Yuan]{papachristou2024network}
Marios Papachristou and Yuan Yuan.
\newblock Network formation and dynamics among multi-llms, 2024.

\bibitem[Park et~al.(2023)Park, O'Brien, Cai, Morris, Liang, and Bernstein]{park2023generative}
Joon~Sung Park, Joseph O'Brien, Carrie~Jun Cai, Meredith~Ringel Morris, Percy Liang, and Michael~S Bernstein.
\newblock Generative agents: Interactive simulacra of human behavior.
\newblock In \emph{Proceedings of the 36th Annual ACM Symposium on User Interface Software and Technology}, pp.\  1--22, 2023.

\bibitem[Perez et~al.(2024)Perez, Léger, Ovando-Tellez, Foulon, Dussauld, Oudeyer, and Moulin-Frier]{perez2024cultural}
Jérémy Perez, Corentin Léger, Marcela Ovando-Tellez, Chris Foulon, Joan Dussauld, Pierre-Yves Oudeyer, and Clément Moulin-Frier.
\newblock Cultural evolution in populations of large language models, 2024.

\bibitem[Radford et~al.(2023)Radford, Kim, Xu, Brockman, McLeavey, and Sutskever]{radford2023robust}
Alec Radford, Jong~Wook Kim, Tao Xu, Greg Brockman, Christine McLeavey, and Ilya Sutskever.
\newblock Robust speech recognition via large-scale weak supervision.
\newblock In \emph{International Conference on Machine Learning}, pp.\  28492--28518. PMLR, 2023.

\bibitem[Reimers \& Gurevych(2019)Reimers and Gurevych]{reimers2019sentence}
Nils Reimers and Iryna Gurevych.
\newblock Sentence-bert: Sentence embeddings using siamese bert-networks.
\newblock \emph{arXiv preprint arXiv:1908.10084}, 2019.

\bibitem[Risch \& Krestel(2019)Risch and Krestel]{risch2019domain}
Julian Risch and Ralf Krestel.
\newblock Domain-specific word embeddings for patent classification.
\newblock \emph{Data Technologies and Applications}, 53\penalty0 (1):\penalty0 108--122, 2019.

\bibitem[Shumailov et~al.(2024)Shumailov, Shumaylov, Zhao, Gal, Papernot, and Anderson]{shumailov2024curse}
Ilia Shumailov, Zakhar Shumaylov, Yiren Zhao, Yarin Gal, Nicolas Papernot, and Ross Anderson.
\newblock The curse of recursion: Training on generated data makes models forget, 2024.

\bibitem[Steels(1990)]{steels1990cooperation}
Luc Steels.
\newblock Cooperation between distributed agents through self-orcamsation.
\newblock In \emph{Proceedings of the first European workshop on modelling autonomous agents in a multi-agent world}. Citeseer, 1990.

\bibitem[Thrun(1995)]{thrun1995learning}
Sebastian Thrun.
\newblock Is learning the n-th thing any easier than learning the first?
\newblock \emph{Advances in neural information processing systems}, 8, 1995.

\bibitem[Thrun(1998)]{thrun1998lifelong}
Sebastian Thrun.
\newblock Lifelong learning algorithms.
\newblock In \emph{Learning to learn}, pp.\  181--209. Springer, 1998.

\bibitem[Torgerson(1952)]{torgerson1952multidimensional}
Warren~S Torgerson.
\newblock Multidimensional scaling: I. theory and method.
\newblock \emph{Psychometrika}, 17\penalty0 (4):\penalty0 401--419, 1952.

\bibitem[Touvron et~al.(2023)Touvron, Martin, Stone, Albert, Almahairi, Babaei, Bashlykov, Batra, Bhargava, Bhosale, et~al.]{touvron2023llama}
Hugo Touvron, Louis Martin, Kevin Stone, Peter Albert, Amjad Almahairi, Yasmine Babaei, Nikolay Bashlykov, Soumya Batra, Prajjwal Bhargava, Shruti Bhosale, et~al.
\newblock Llama 2: Open foundation and fine-tuned chat models.
\newblock \emph{arXiv preprint arXiv:2307.09288}, 2023.

\bibitem[Vogelstein et~al.(2020)Vogelstein, Helm, Mehta, Dey, Yang, Tower, LeVine, Larson, White, and Priebe]{vogelstein2020general}
Joshua~T Vogelstein, Hayden~S Helm, Ronak~D Mehta, Jayanta Dey, Weiwei Yang, Bryan Tower, Will LeVine, Jonathan Larson, Chris White, and Carey~E Priebe.
\newblock A general approach to progressive learning.
\newblock \emph{Preprint at https://arxiv. org/abs/2004.12908}, 2020.

\bibitem[Wagner et~al.(2003)Wagner, Reggia, Uriagereka, and Wilkinson]{wagner2003progress}
Kyle Wagner, James~A Reggia, Juan Uriagereka, and Gerald~S Wilkinson.
\newblock Progress in the simulation of emergent communication and language.
\newblock \emph{Adaptive Behavior}, 11\penalty0 (1):\penalty0 37--69, 2003.

\bibitem[Wang et~al.(2020{\natexlab{a}})Wang, Zhang, Wang, Ma, and Liu]{9044329}
Fangxin Wang, Miao Zhang, Xiangxiang Wang, Xiaoqiang Ma, and Jiangchuan Liu.
\newblock Deep learning for edge computing applications: A state-of-the-art survey.
\newblock \emph{IEEE Access}, 8:\penalty0 58322--58336, 2020{\natexlab{a}}.
\newblock \doi{10.1109/ACCESS.2020.2982411}.

\bibitem[Wang et~al.(2020{\natexlab{b}})Wang, Wei, Dong, Bao, Yang, and Zhou]{wang2020minilm}
Wenhui Wang, Furu Wei, Li~Dong, Hangbo Bao, Nan Yang, and Ming Zhou.
\newblock Minilm: Deep self-attention distillation for task-agnostic compression of pre-trained transformers.
\newblock \emph{Advances in Neural Information Processing Systems}, 33:\penalty0 5776--5788, 2020{\natexlab{b}}.

\bibitem[Wolf et~al.(2020)Wolf, Debut, Sanh, Chaumond, Delangue, Moi, Cistac, Rault, Louf, Funtowicz, Davison, Shleifer, von Platen, Ma, Jernite, Plu, Xu, Scao, Gugger, Drame, Lhoest, and Rush]{wolf2020huggingfaces}
Thomas Wolf, Lysandre Debut, Victor Sanh, Julien Chaumond, Clement Delangue, Anthony Moi, Pierric Cistac, Tim Rault, Rémi Louf, Morgan Funtowicz, Joe Davison, Sam Shleifer, Patrick von Platen, Clara Ma, Yacine Jernite, Julien Plu, Canwen Xu, Teven~Le Scao, Sylvain Gugger, Mariama Drame, Quentin Lhoest, and Alexander~M. Rush.
\newblock Huggingface's transformers: State-of-the-art natural language processing, 2020.

\bibitem[Zamir et~al.(2018)Zamir, Sax, Shen, Guibas, Malik, and Savarese]{zamir2018taskonomy}
Amir~R Zamir, Alexander Sax, William Shen, Leonidas~J Guibas, Jitendra Malik, and Silvio Savarese.
\newblock Taskonomy: Disentangling task transfer learning.
\newblock In \emph{Proceedings of the IEEE conference on computer vision and pattern recognition}, pp.\  3712--3722, 2018.

\bibitem[Zhang et~al.(2015)Zhang, Zhao, and LeCun]{zhang2015character}
Xiang Zhang, Junbo Zhao, and Yann LeCun.
\newblock Character-level convolutional networks for text classification.
\newblock \emph{Advances in neural information processing systems}, 28, 2015.

\bibitem[Zuzul et~al.(2023)Zuzul, Pahnke, Larson, Bourke, Caurvina, Shah, Amini, Weston, Park, Vogelstein, White, and Priebe]{zuzul2023dynamic}
Tiona Zuzul, Emily~Cox Pahnke, Jonathan Larson, Patrick Bourke, Nicholas Caurvina, Neha~Parikh Shah, Fereshteh Amini, Jeffrey Weston, Youngser Park, Joshua Vogelstein, Christopher White, and Carey~E. Priebe.
\newblock Dynamic silos: Increased modularity in intra-organizational communication networks during the covid-19 pandemic, 2023.

\end{thebibliography}

\clearpage

\appendix

\section{Instruction-tuning Pythia-410m-deduped}
\label{app:insruction-tuning}

The base model that we used in the case studies in Section \ref{sec:simulations} was an instruction-tuned version of the 410 million parameter model from the Pythia suite \citep{biderman2023pythia}. 
For instruction-tuning, we added three special tokens to its tokenizer's vocabulary, ``\#\#\# End", ``\#\#\# Instruction:", and ``\#\#\# Response:", and fine-tuned the model with a subset of Databricks' Dolly 15k \citep{DatabricksBlog2023DollyV2}.
Each datum consists of an instruction, context, response, and category.
We kept only data in the Open QA, Brainstorm, General QA, and Creative Writing categories and that had a response length less than 100 characters.
This filtering left us with 1559 instruction-response pairs.
We formatted a particular example as follows:
\begin{align*}
    &\text{\#\#\# Instruction: \{instruction\}} \\
    &\text{\#\#\# Response: \{response\}} \\
    &\text{\#\#\# End}
\end{align*}

We fine-tuned the model on the formatted data using Adam with a learning rate of $ 5 \times 10^{-5} $ and a batch size of 8 for 10 epochs. The final cross-entropy loss on the training data was $ \approx 0.26 $.

\section{Case-study specific fine-tuning}
\label{app:fine-tuning}
For each of the case studies we further fine-tuned the instruction-tuned base model to promote response variation. 
For this, we used the data from the Yahoo! Answers (YA) dataset introduced in \citep{zhang2015character}, where each datum consists of a topic, a question title, question content, a list of answers, and a best answer.
Given data from a particular topic, we further filtered the data by considering only examples with best answers less than 200 characters, with best answers that contained only a single sentence, and with question titles that contained only a single question.
We formatted data from YA as follows:
\begin{align*}
    &\text{\#\#\# Instruction: \{question title\}} \\
    &\text{\#\#\# Response: \{best answer\}} \\
    &\text{\#\#\# End}
\end{align*}

Unless otherwise specified, fine-tuning is done using Adam with a learning rate of $ 5 \times 10^{-5} $. The initial models were trained for 3 epochs. The model updates after an interaction consisted of only a single epoch with a learning rate of $ 10^{-5} $.

\subsection{Case Study 1: Stochastically Equivalent Models}

For case study 1, we randomly selected 400 examples with the topic ``Society \& Culture" that we used as both the evaluation set in the experiment and as a pool of data used for further sampling.
In particular, we randomly sampled 200 samples from the set of 400 25 times and used the 25 subsets as fine-tuning data for different ``stochastically equivalent" models.

\subsection{Case Studies 2 \& 3: Two classes}

For case studies 2 \& 3, we considered filtered data from topics ``Society \& Culture" and ``Science \& Mathematics". For each topic we randomly sampled 1000 examples 10 times to use for fine-tuning.

For case study 2, we randomly selected a single model fine-tuned on ``Science \& Mathematics" to be the adversarial model. This model was the adversarial model for all system instances. We then randomly selected 5 models fine-tuned on ``Society \& Culture" data to be non-adversarial models. The non-adversarial models changed with each system instance.

For case study 3, we randomly selected 5 models from each class for every system instance.


\end{document}